# A Comprehensive Survey of Belief Rule Base (BRB) Hybrid Expert system: Bridging Decision Science and Professional Services

Karim Derrick, University of Manchester

## 1 Abstract

The Belief Rule Base (BRB) system that adopts a hybrid approach integrating the precision of expert systems with the adaptability of data-driven models. Characterized by its use of if-then rules to accommodate various types of uncertainty through belief degrees, BRB adeptly handles fuzziness, randomness, and ignorance. This semi-quantitative tool excels in processing both numerical data and linguistic knowledge from diverse sources, making it as an indispensable resource in modelling complex nonlinear systems. Notably, BRB's transparent, white-box nature ensures accessibility and clarity for decision-makers and stakeholders, further enhancing its applicability. With its growing adoption in fields ranging from decision-making and reliability evaluation in network security and fault diagnosis, this study aims to explore the evolution and the multifaceted applications of BRB. By analysing its development across different domains, we highlight BRB's potential to revolutionize sectors traditionally resistant to technological disruption, in particular insurance and law.

## 2 Introduction

The objective of this study is to explore the degree to which the increasingly popular use of the decision science methodology Belief Rule Base (BRB) has application in professional services. BRB is a decision support algorithm used in modelling complex nonlinear systems that has demonstrated excellent application in engineering contexts where uncertainty is a significant factor. Adoption of BRB in engineering contexts has been growing exponentially in recent years; this study investigates the evolution of BRB conceptually and in application to demonstrate the potential and novelty of its exploration in professional services and in particular law and insurance.

Professional services in the UK generate over £500 billion in revenue, represent 20% of all business but have to date largely escaped disruption from technology. Uncertainty, incompleteness, and vagueness are all features of insurance legal datasets. This has implications for the notion of human professionals: “Like a good measuring instrument, an expert judge must be both discriminating and consistent” (Weiss and Shanteau 2003). Recent work from Daniel Kahneman has highlighted an absence of reliability in the judgement of insurance professionals and estimates the cost impact of inconsistency in human decision making to be in the hundreds of millions for a single US insurance company alone (Daniel Kahneman 2021) thus making any technology that can improve decisioning in the professional services of huge potential benefit.

Numerous studies have demonstrated that professionals often lack consistency in life changing decisioning and for decades now it has been shown that algorithms can exceed professional performance. Meehl’s seminal work demonstrates that statistical models are almost always more accurate than clinicians (Meehl 1954). Goldberg’s work showed that a statistical model of an individual clinician was more accurate than the clinician modelled (Goldberg 1976).

The contribution of this paper is to analyse the development of BRB across multiple domains and to consider the potential application across the professional and financial services.

# 3 Review Method

The methodology used to explore the BRB literature is based on the protocols and procedures elaborated by Sinkovics to ensure analytical rigor and also to ensure the reader's ability to recreate the author's investigation (Sinkovics 2016).

A key word search was conducted in the Web of Science using the following query:

AB=("belief rule base*" or "belief rules base*") or TI=("belief rule base*" or "belief rules base*")

The initial search retrieved 349 articles in January 2023. An additional search in January 2024 produced a further 63 papers. The resulting inventory was then organised by chronology to enable the evolution of BRB to be detailed, by domain where the article detailed a practical application of BRB to determine the extent to which the approach has been applied to Professional Services, by methodology where the article detailed and tested a development or refinement of the original BRB concept and by reference to reliability or explainability, both important concepts in respect of any potential application of BRB in the Professional Services.

The analysis followed two steps in line with Sinkovics (2016). Firstly the free VOSviewer (www.vosviewer.com) was used to perform a bibliographic analysis where the relatedness of each article is determined based on the number of references that are shared (Van Eck and Waltman 2014). See Figure 1 below.

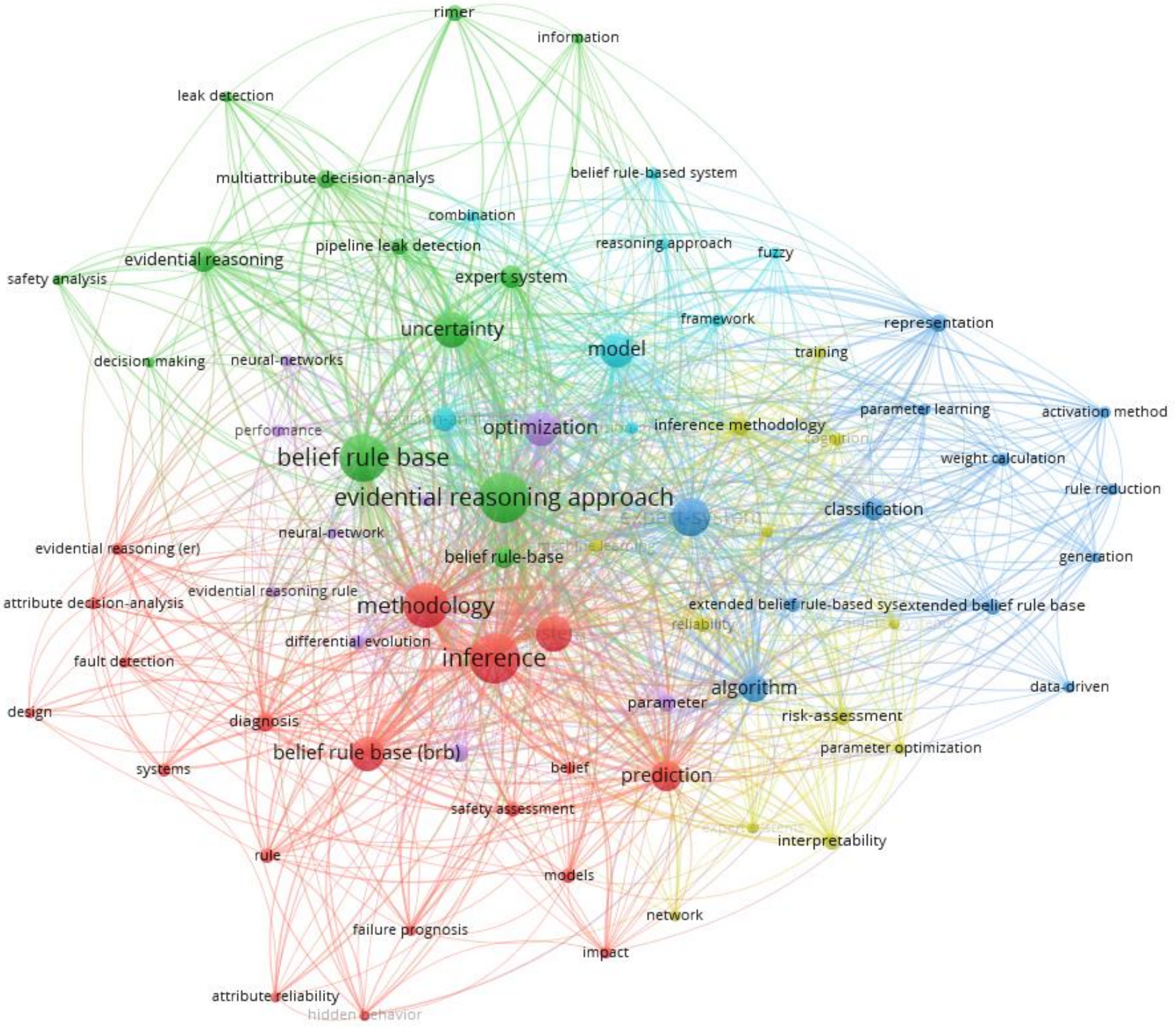

*Figure 1*

In the second step the abstracts and bibliographic data was loaded into Nvivo having first imported all the PDFs into EndNote. A thematic analysis was then conducted using NVivo, conducting analysis over how the focuses and themes of BRB research develop over time.

# 4 Findings

## 4.1 Chronology

### 4.1.1 RIMER

The original BRB paper, and the most cited (Yang, Liu et al. 2006), introduces a generic rule-base inference methodology using the evidential reasoning approach, otherwise called RIMER. Evidential Reasoning is a development of Dempster-Shafer Theory, a mathematical framework for modelling uncertainty about knowledge that addresses issues with the original Rule of Combination that has been shown to lead to counterintuitive results where there is conflict between different evidence sources (Shafer 1976, Yang and Xu 2002). The BRB approach, combined with evidential reasoning, is designed specifically to address the issue of uncertainty in human decision making, driven by the vagueness, imprecision and incompleteness that is feature of so much human decision making. All of these are features of decision making in respect of all professional services. Yang's paper proposes a new knowledge representation scheme that enables:

- a belief-structure where belief degrees are embedded into all possible consequents of a rule enabling RIMER to capture uncertainty, incompleteness and non-linear causal relationships.
- inputs to antecedent attributes in rules to be transformed into belief distributions on the referential values of antecedents. This distribution describes the degree to which each antecedent is activated.
- the weights of both attributes and rules to be considered.
- inference on rule bases to be implemented using the evidential reasoning (ER) approach.
- inference from hierarchical rule bases.

The paper ends noting the problem of consistency of the rule base when it is generated by expert knowledge, suggesting that further investigation is required for an issue that will have particular importance in the fields of law and insurance. As will be seen, a lot work has been done on this issue subsequently.

Offline optimisation methods for RIMER are introduced by Yang, Liu et al. (2007) securing the approaches' position as a hybrid approach, combining both expertise and data.

A seminal pipeline leak detection paper is introduced in Xu, Liu et al. (2007) that uses a pipeline data set to demonstrate the efficacy of the BRB approach that is then subsequently used by many studies to demonstrate incremental improvements to the core BRB approach.

### 4.1.2 Extended Belief Rule Base

Despite the focus of the original RIMER paper on uncertainty, the original approach does not take account of vagueness or fuzzy uncertainty of *consequents* in the IF-THEN rule, instead assuming that each consequent is independent and crisp and thus limiting the applicability of the approach (Liu, Martinez et al. 2008). To overcome this RIMER is extended to the case of fuzzy consequents where each consequent can be defined in fuzzy linguistic terms because of their “vagueness and inexactness”. This is extension is entitled the Extended Belief Rule Base and represents a significant milestone in the evolution of BRB.

In Liu’s example, assessment grades “low” and “very low” are not naturally expressed as distinctive crisp sets, but instead as dependent fuzzy sets: “In other words, the intersection of the two fuzzy sets may not be empty.”

A new online recursive algorithm for training BRB models (Zhou, Hu et al. 2009) allows RIMER parameters to be updated as soon as new information becomes available based on the recursive-expectation-maximization algorithm. This is later updated with a Bayesian reasoning approach (Zhou, Hu et al. 2011).

As an extension to the original approach a local training model is proposed in 2011 to optimize all the parameters in a BRB system including not only belief degrees , rule weights and attribute weights as considered in the original local training model (Yang, Liu et al. 2007), but to also now include the referential values of antecedent attributes (Chen, Yang et al. 2011).

The issue of consistency in rules is also considered in Liu, Martinez et al. (2011) noting that real world data will often feature inconsistency and that the BRB optimisation strategies originally proposed by Yang, Liu et al. (2007)  did not take this into account. The new optimisation methodology includes the measurement of BRB inconsistency which can then be incorporated into the optimization algorithm to minimise inconsistency.

Chen, Yang et al. (2013) investigate the underlying inference mechanisms that enable BRB system to have superior approximation performance and prove that BRB systems can approximate any continuous function on a compact set.

Calzada, Liu et al. (2015) specifically address incompleteness and inconsistency with their paper, introducing dynamic rule activation (DRA), a method for rule activation that searches to balance inconsistency and incompleteness.

The Extended Belief Rule base (EBRB) is developed further to enable a new simple procedure for learning rules based on numerical data (Liu, Martinez et al. 2013). The method is shown to outperform the expert derived models that were used in the original pipeline leak detection paper (Xu, Liu et al. 2007). The consistency of the generated rule base is also considered acknowledging that inconsistency generally exists in the process of knowledge representation and acquisition and that rules generated by inconsistent data will have serious consequences. We know that this is an issue particularly prevalent in the Professional Services (Daniel Kahneman 2021). The approach includes measurement of EBRB inconsistency, which is then incorporated into rule weights, improving the performance of the EBRB system by reducing the inconsistency via adjusted rule weights.

Significant papers include the use of data envelopment analysis to reduce the number of rules in ERBB systems, evaluating the efficiency of each rule in an extended belief-rule-based (Yang, Wang et al. 2017).

A branch of BRB research begins in 2017 that challenges the conjunctive assumption that the original BRB approach features and develops a disjunctive approach (Xiong, Chen et al. 2017, Yang, Wang et al. 2017). This new approach addresses the issue that many papers cite when too many attributes create a significant computational overhead often referred to in the literature as the combinatorial explosion problem. Another 10 papers go on to explore the potential for a disjunctive assumption.

Following the development of data only driven applications of BRB, several studies look at classification applications using common data sets to explore and iteratively develop the efficiency of the approach. The most cited classification paper demonstrates the use of BRB with evidential reasoning as the inference engine and the differential evolution (DE) algorithm over the commonly used iris, wine, glass, cancer and pima datasets to identify the fittest parameters, including the referenced values of the antecedent attributes, the weights of the rules and the beliefs of the degrees in the result (Chang, Zhou et al. 2016).

It is not until 2018 that papers begin to emerge to consider the interpretability of trained BRBs, ironic given that the interpretability of BRB is often hailed as a positive quality in contrast to black box neural network approaches (Sachan, Yang et al. 2020, Cao, Zhou et al. 2021). With just one paper in 2020 on explainability, this had grown to seven papers in 2023 perhaps reflecting the growing concern across the wider AI community for explainability. Yin, Jia et al. (2023) introduce the concept of calculating expert credibility recognising that BRB often assumes expert knowledge to be reliable.

## 4.2 BRB Development

Over half of all the papers on BRB look to develop the original Yang, Liu et al. (2006) proposal. Of those the majority explore optimisation. Initially these are focussed on optimisation of parameters, then of structure, then of parameters and structure. Later on, attention is drawn to the issues of reliability and explainability, issues which have particular relevance in both the legal and insurance knowledge and representation.

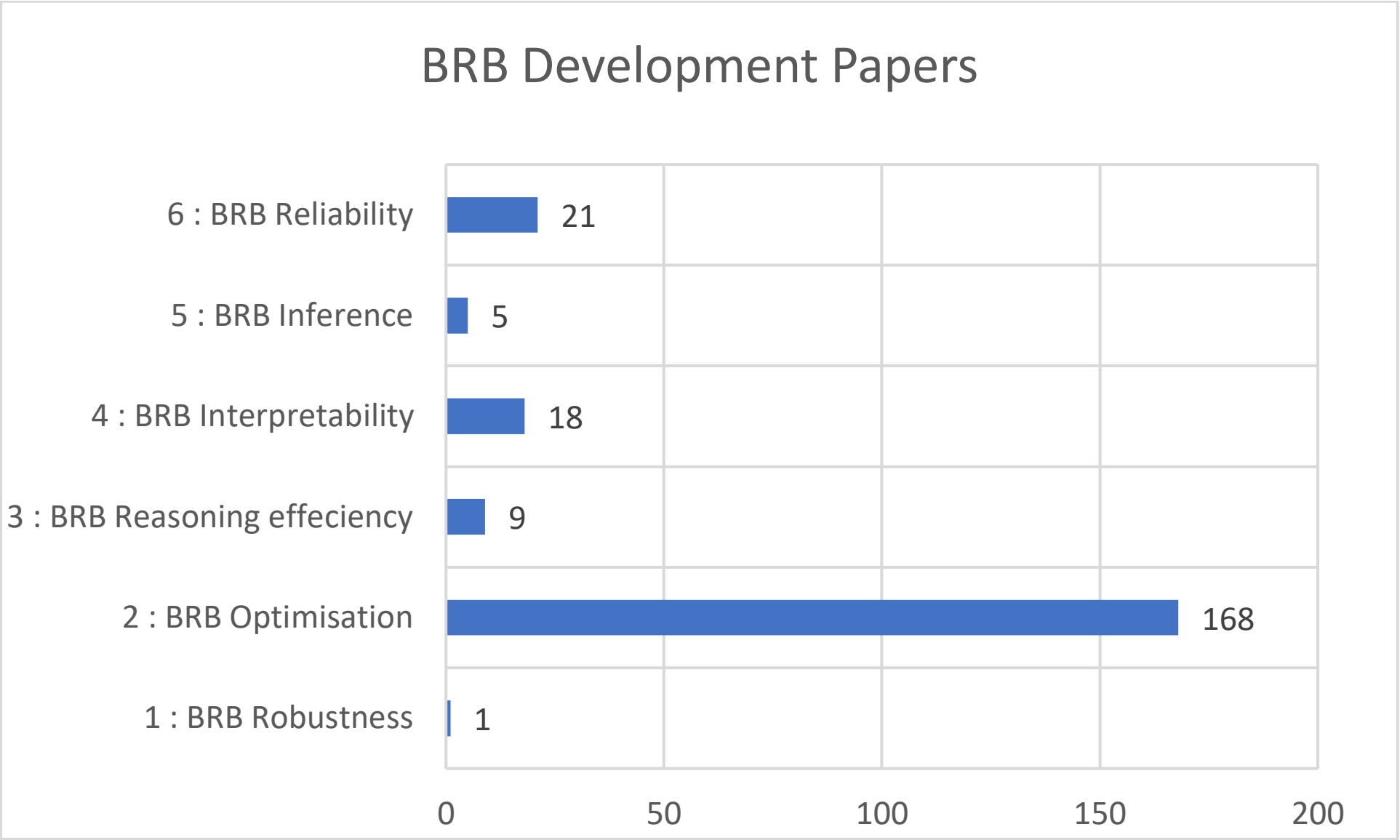

*Figure 2*

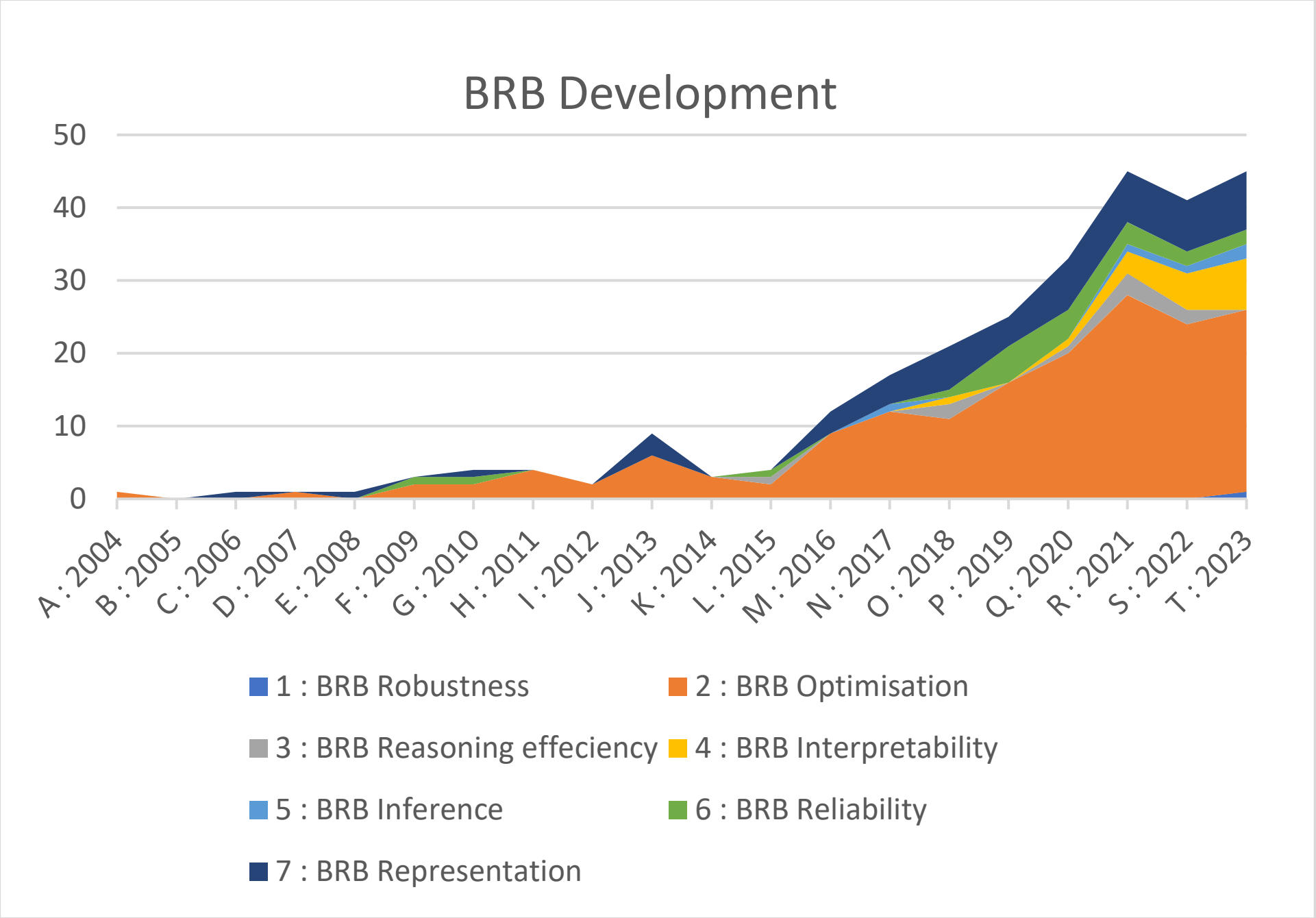

*Figure 3*

## 4.3 Optimisation

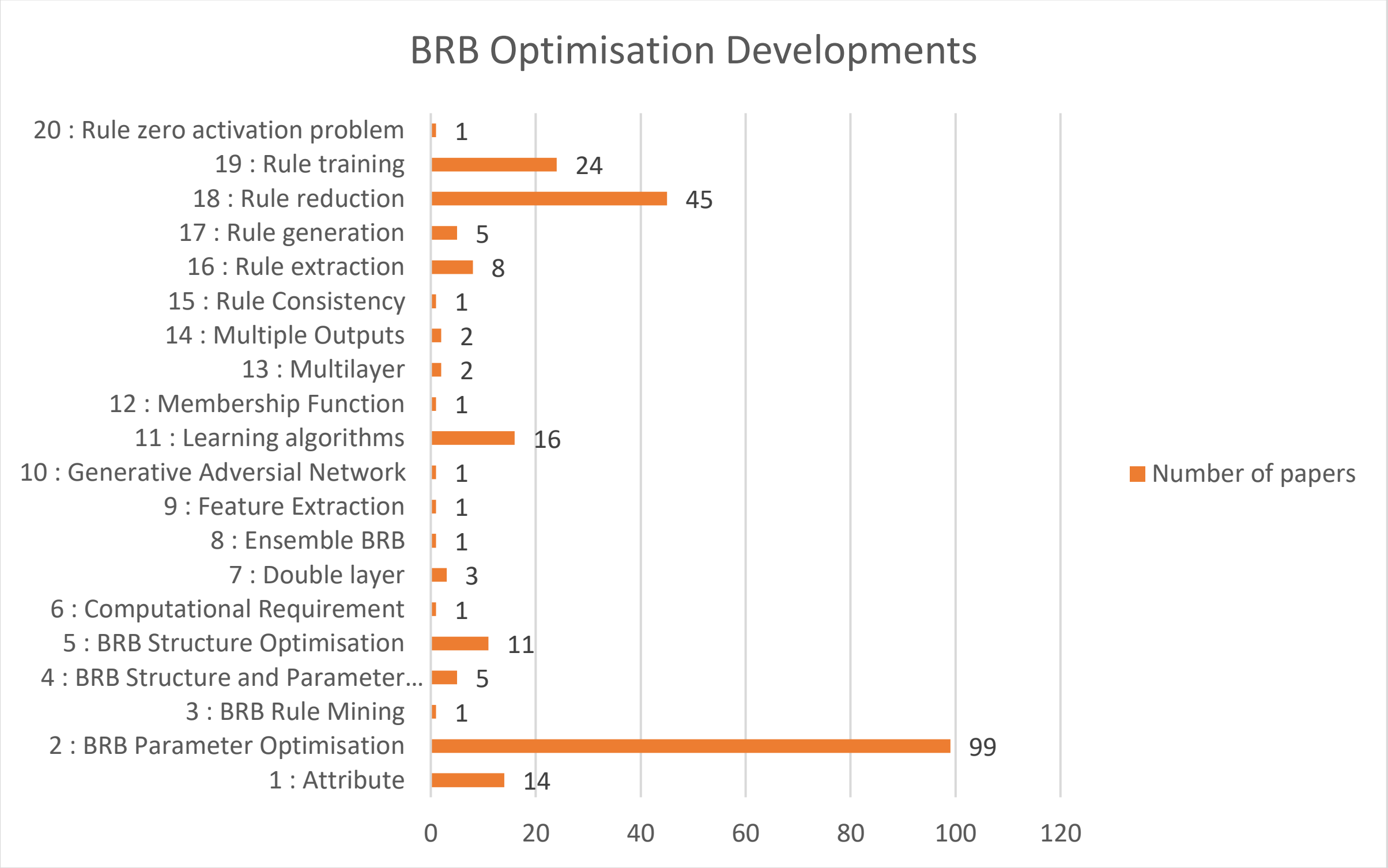

*Figure 4*

The first parameter optimsation paper for RIMER, Yang, Liu et al. (2007) developed the original RIMER concept for BRB.

Chen, Yang et al. (2011) introduced a general adaptive parameter optimization model based on a heuristic training data selection scheme based on the Optimisation Toolbox in Matlab.

Zhou proposed the first online parameter-learning approach based on recursive-expectation-maximization (Zhou, Hu et al. 2011).

The Differential Evolution (DE) algorithm is introduced as an optimisation engine when optimising the parameters in a BRB with another 15 papers adopting the technique in subsequent years (Chang, Sun et al. 2015, Chang, Zhou et al. 2016).

Eventually it became clear that training BRB parameters alone is not sufficient, adjusting the BRB structure is also necessary. The effectiveness of parameter learning it itself a function of the structure of the BRB. Methods to determines the optimal structure in a BRB are developed, which optimise the number of parameters rather than the parameters themselves. Too many parameters may result in over-fitting. Too few may result in under-fitting. Thus, the dimensionality reduction-based structure learning method (Chang, Zhou et al. 2013) and the dynamic rule adjustment approach are all introduced to specifically target structure (Wang, Yang et al. 2016).

Wang, Yang et al. (2009) used Principle Component Analysis (PCA) to reduce the number of antecedent attributes in BRB before parameter optimisation methods are used to optimise parameters and this is successfully applied in consumer prediction contexts. Subsequent studies demonstrated similar success (Yang, Wang et al. 2012, Yang, Fu et al. 2016) with 29 papers mentioning some use of PCA.

Chang, Zhou et al. (2018) proposed a bi-level optimization method with Akaike Information Criterion (AIC)-based objectives to achieve parameter and structure optimization of the BRB simultaneously.

Yang, Wang et al. (2018) also introduced a method to optimise both parameters and structure. Density and error analyses are introduced to construct the BRB structure.

### 4.3.1 Intervals

A relatively new branch of BRB development has emerged in respect of interval belief rule base. This approach models a process using interval addition combination rules to address the issue of rule explosion in complex systems in combination with an optimisation process that imposes interpretable constrains to retain interpretability (He, Cheng et al. 2023). The original BRB approach gives rise to rule explosions because of Cartesian product combination rules. Two attributes with four referential points produce 16 rules. 3 attributes with 4 referential points generate 64 rules. This new approach uses interval addition combined with an adapted rule activation methodology.

### 4.3.2 Language models and natural language processing

Large language models have exploded onto the machine intelligence scene in recent years following the original transformer paper (Devlin, Chang et al. 2018) and then with the step change launch of Open AI’s GPT 3.5 in late 2022. Language models have been developed to predict the next word in a given text, or the next word given a text as context; as such they are a significant development in the natural language processing field. The basic models are trained on very large corpuses of text and are can then be fine-tuned on domain or task specific texts. Fine-tuning language models allows the model weights to be adjusted for down-stream natural language tasks. Unsurprisingly, researchers have begun to explore the potential opportunity for combining capabilities of large language models with the benefits of the BRB approach. (Kang, Xiao et al. 2023) develop a method for few-shot health condition estimation that uses generative techniques to generate training data. (Chen, Zhou et al. 2023) use language processing to extract belief rules from text. In consideration of the application of BRB to insurance and law, the ability to extract attribute values or even rules from case files and correspondents is an attractive possibility with most historical data relating to law and insurance dominated by unstructured text.

## 4.4 Domains

The focus of early BRB papers has largely been theoretical with development often focused around a narrow set of applications to enable comparison of optimisation techniques, thus the original focus on leak detection and classification. Recent years though have seen an explosion of literature testing BRB in specific domains as illustrated by Figure 5 and Figure 6.

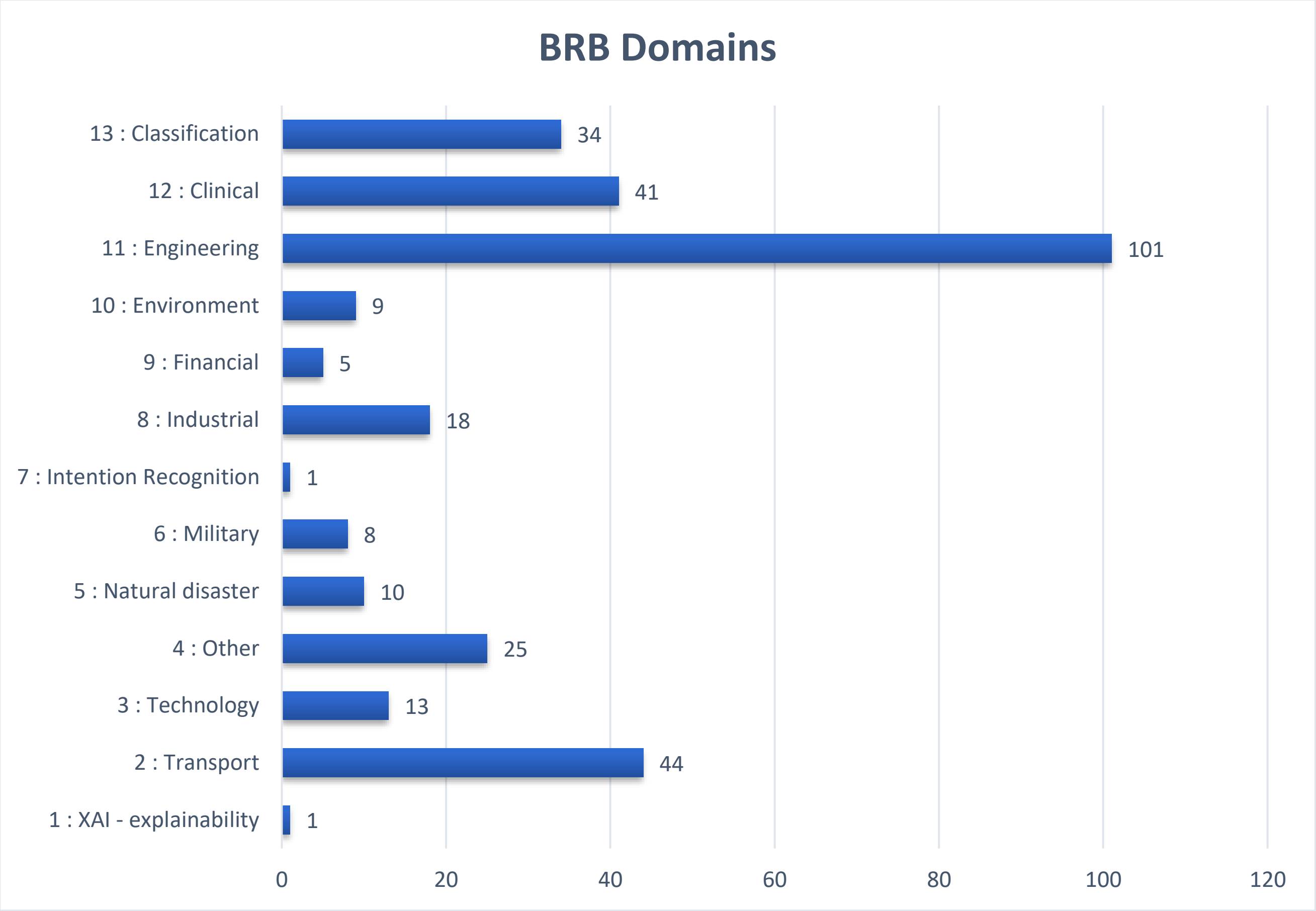

*Figure 5*

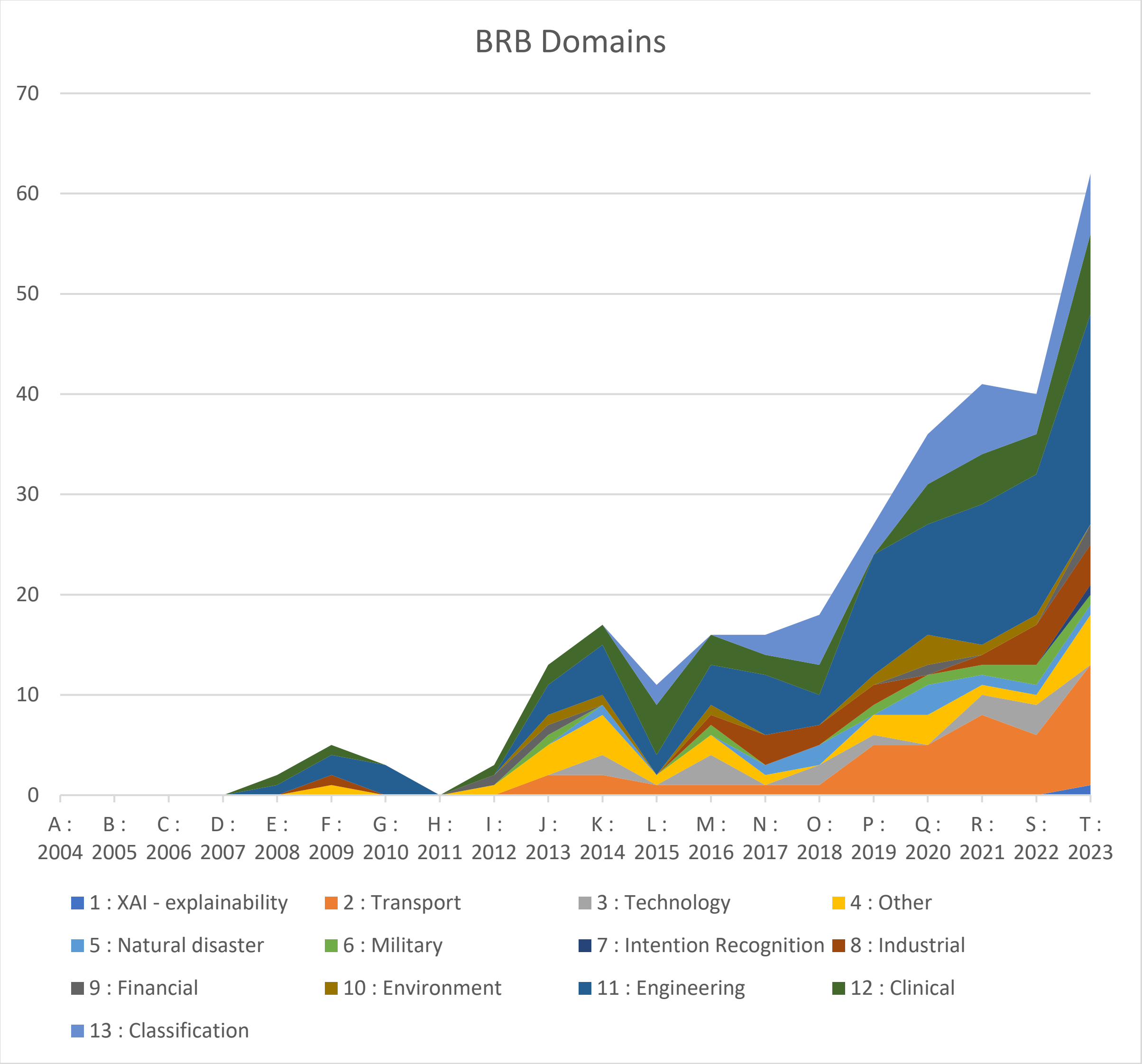

*Figure 6*

### 4.4.1 Engineering

#### *4.4.1.1 Fault Diagnosis*

The original RIMER paper used an example from radiological waste disposal that was used as the basis for developing the use of fuzzy logic in expert systems.(Hodges, Bridges et al. 1996)

Xu, Liu et al use pipeline fault diagnosis in the 2007 optimsiation paper. (Xu, Liu et al. 2007).

33% of all the BRB papers reviewed have an engineering focus with 50% of those with fault diagnosis as their focus.

The first paper to look specifically at failure prognosis generally in 2010 combines RIMER (Chen, Yang et al. 2011) BRB with a hidden Markov model that also includes a recursive algorithm for online updating (Zhou, Hu et al. 2010).

#### 4.4.1.2 Industrial

22 papers with an industrial focus range from petrochecmical production (Ruan, Carchon et al. 2009, Zou, Yang et al. 2016), industrial control systems (Hossain, Rahaman et al. 2017, Islam, Ruci et al. 2019), data center management (Xu, Ma et al. 2018) to nuclear safeguards (Ruan, Carchon et al. 2009).

#### 4.4.1.3 Clinical

The first clinical support system using BRB is proposed in 2012 and is designed to triage patients reporting cardiac chest pain addressing many of the shortcomings of the MYCIN (Shortliffe 1976) clinical support system in Kong, Xu et al. (2012), consistently demonstrating an improvement over manual diagnosis and cited 76 times since then.

40 papers, 13% of the total have a clinical diagnosis focus ranging from chest pain, to cancer (Hossain, Hasan et al. 2015), to COVID (Ahmed, Hossain et al. 2021) in recent times. Clinical diagnosis is perhaps the closest domain that has had focus from BRB research to the domains of legal and insurance and is the clearest indication of the potential for application in those domains.

#### 4.4.1.4 Transport

The application of BRB in transport is first explored in 2013 with a paper that looks at real-time fault diagnosis of aircraft navigation systems (Xie, He et al. 2022).

Most recently drone detection has featured reflecting growing security concerns across the world and the growth in the importance of UAV vehicles in modern warfare (Zhao, Wang et al. 2013). 8 papers across the total cohort have a military focus, most of those with a target classification focus, from carrier groups (Jiao, Pan et al. 2013) to aerial targets. (Liu, Zhou et al. 2020)

Over a quarter of the transport themed papers are from the last year (2023) clearly representing growth in application in this domain.

#### 4.4.1.5 Non-engineering

Outside of engineering, clinical and industrial applications there are fewer applications of BRB to explore and only one within the insurance or legal domains.

Perhaps the most unusual is BRB in relation to the design of a lemonade drink product using real data provided by a sensory product manufacturer in the UK demonstrating that the methodology can be used to predict consumer preferences and to set optimal target values for product quality improvement (Yang, Wang et al. 2012).

10 papers consider natural disaster risk including 5 on flood (Hridoy, Ul Islam et al. 2017, Monrat, Ul Islam et al. 2018, Ul Islam, Hossain et al. 2020) and 2 on pollution risk(Kabir, Islam et al. 2020, Ye, Yang et al. 2020).

Sentiment analysis is the subject of a paper that explores how BRB can be used to categorise human sentiment from analysis of images (Zisad, Chowdhury et al. 2021).

Multiple papers explore the use of BRB in the financial sector explore the potential for using BRB to affect stock trading decisioning. With 3 papers in the last year, the need for ensuring explainability in the financial services may well be driving more interest in this area (Stoia 2013, Hossain, Hossain et al. 2022, Chen, Liu et al. 2023, Yin, Zhang et al. 2023) .

In 2013 with a BRB approach for affecting a failure and effects analysis using evidential reasoning and BRB. (Liu, Liu et al. 2013).

## 4.5 Interpretability

BRB has been favoured in part for it's non-black box features. However, increasingly in recent times use of the Extended Belief Rule Base in data only contexts have meant that a growing number of papers have started to consider the interpretability of BRB.

Cao, Zhou et al. (2021) establish interpretability criteria for BRB systems noting that to date all research effort has been focussed on accuracy.

You, Sun et al. (2022) explore the trade-off between interpretability and accuracy, designing a framework for measuring interpretability which is then used as a basis for an optimisation method which uses the differential evolution (DE) algorithm to ensure the interpretability and accuracy of BRB systems simultaneously, especially where interpretability is required.

Very recently Han, He et al. (2023) explore the possibility of the whale optimisation algorithm to enhance the interpretability of the BRB approach. The Whale Optimisation Algorithm was developed based on the foraging behaviour of humpback whales, and does not require gradient information. The algorithm uses a unique bubble net hunting method to search, moving from the outermost layer of tha spiral trajectory to the centre point.

## 4.6 Reliability

The original RIMER paper (Yang, Liu et al. 2006) makes reference to the problem of noise in the expert knowledge that is represented: "Consistency between generated rules with the intuition and common sense of human beings needs to be further investigated."

Subsequent papers look explicitly at the issue of rule consistency and explore approaches to optimising in relation to consistency. The most cited proposes a dynamic rule activation (DRA) method to select activated rules in a dynamic way optimising the balance between the incompleteness and inconsistency in the rule-base generated from sample data to achieve a better performance. (Calzada, Liu et al. 2015)

Given the significant inconsistency issue that has been exposed by Daniel Kahneman (2021) the opportunity to optimise consistency is attractive for the legal and insurance domains.

## 4.7 Influential Papers and Authors

Whilst Jian-Bo Yang from University of Manchester UK wrote the original BRB paper (Yang, Liu et al. 2006) and is the most cited in respect of BRB, many others have gone on to write many more. Lei-Lei Chang, affiliated to the National University of Defense Technology, Changsha, China has been the most prolific. Longhao Yang of Fuzhou University, Fuzhou, China is a close second. Zhijie Zhou from Guangdong University of Technology, China has led less papers but has been cited substantially more. Mohammad Shahadat Hossain Department of Computer Science and Engineering, University of Chittagong, Bangladesh with 10 papers has been prolific in his exploration of BRB clinical diagnosis contexts.

Dong-Ling Xu of University of Manchester has led only 1 paper but it has been cited substantially, introducing the pipeline fault detection data set that many papers subsequently test their own methods against.

Dr Yu-wang Chen has contributed only 5 papers but has been cited significantly by his contribution to evidencing the accuracy and scope of BRB in nonlinear complex contexts.

Alberto Calzada of European University Cyprus has been active in BRB contributing the Dynamic Rule Activtation method for Extended Belief Rule bases which has inspired an entire branch of paper.

As is shown by Figure 9 China dominates the table of countries producing the most BRB papers with the UK and the University of Manchester a distant second.

| Lead Author | Number articles | Total citations |
|---|---|---|
| Chang, L. L. | 20 | 485 |
| Yang, L. H. | 20 | 333 |
| Zhou, Z. J. | 18 | 774 |
| Hossain, M. S. | 10 | 192 |
| Liu, J. | 8 | 218 |
| Cheng, C. | 8 | 95 |
| Feng, Z. C. | 7 | 155 |
| Zhang, B. C. | 7 | 43 |
| Jiao, L. M. | 6 | 107 |
| Xu, X. B. | 6 | 72 |
| Calzada, A. | 6 | 58 |
| Yin, X. J. | 6 | 36 |
| You, Y. Q. | 6 | 29 |
| Chen, Y. W. | 5 | 194 |
| Fu, Y. G. | 5 | 104 |
| Cao, Y. | 5 | 86 |
| He, W. | 5 | 57 |
| Gao, F. | 5 | 34 |
| Yin, X. X. | 5 | 0 |
| Yang, J. B. | 4 | 756 |

*Figure 7*

| Author | Year | Citations |
|---|---|---|
| Yang, J. B.;Liu, J.;Wang, J.;Sii, H. S.;Wang, H. W.; | 2006 | 499 |
| Xu, D. L.;Liu, J.;Yang, J. B.;Liu, G. P.;Wang, J.;Jenkinson, I.;Ren, J.; | 2007 | 203 |
| Yang, J. B.;Liu, J.;Xu, D. L.;Wang, J.;Wang, H. W.; | 2007 | 190 |
| Liu, H. C.;Liu, L.;Lin, Q. L.; | 2013 | 139 |
| Zhou, Z. J.;Hu, C. H.;Xu, D. L.;Chen, M. Y.;Zhou, D. H.; | 2010 | 99 |
| Liu, J.;Martinez, L.;Calzada, A.;Wang, H.; | 2013 | 90 |
| Chang, L. L.;Zhou, Y.;Jiang, J.;Li, M. J.;Zhang, X. H.; | 2013 | 87 |
| Zhou, Z. J.;Hu, C. H.;Yang, J. B.;Xu, D. L.;Zhou, D. H.; | 2009 | 86 |
| Kong, G. L.;Xu, D. L.;Body, R.;Yang, J. B.;Mackway-Jones, K.;Carley, S.; | 2012 | 85 |
| Li, G. L.;Zhou, Z. J.;Hu, C. H.;Chang, L. L.;Zhou, Z. G.;Zhao, F. J.; | 2017 | 85 |
| Xu, X. J.;Yan, X. P.;Sheng, C. X.;Yuan, C. Q.;Xu, D. L.;Yang, J. B.; | 2020 | 83 |
| Chang, L. L.;Zhou, Z. J.;You, Y.;Yang, L. H.;Zhou, Z. G.; | 2016 | 77 |
| Chen, Y. W.;Yang, J. B.;Xu, D. L.;Zhou, Z. J.;Tanga, D. W.; | 2011 | 76 |
| Zhou, Z. J.;Hu, C. H.;Yang, J. B.;Xu, D. L.;Zhou, D. H.; | 2011 | 76 |
| Chowdury, M. S. U.;Bin Emran, T.;Ghosh, S.;Pathak, A.;Alam, M. M.;Absar, N.;Andersson, K.;Hossain, M. S.; | 2019 | 73 |
| Feng, Z. C.;Zhou, Z. J.;Hu, C. H.;Chang, L. L.;Hu, G. Y.;Zhao, F. J.; | 2019 | 73 |
| Zhou, Z. J.;Chang, L. L.;Hu, C. H.;Han, X. X.;Zhou, Z. G.; | 2016 | 71 |
| Chen, Y. W.;Yang, J. B.;Xu, D. L.;Yang, S. L.; | 2013 | 69 |
| Zhou, Z. J.;Hu, G. Y.;Hu, C. H.;Wen, C. L.;Chang, L. L.; | 2021 | 65 |
| Xu, X. J.;Zhao, Z. Z.;Xu, X. B.;Yang, J. B.;Chang, L. L.;Yan, X. P.;Wang, G. D.; | 2020 | 63 |

*Figure 8*

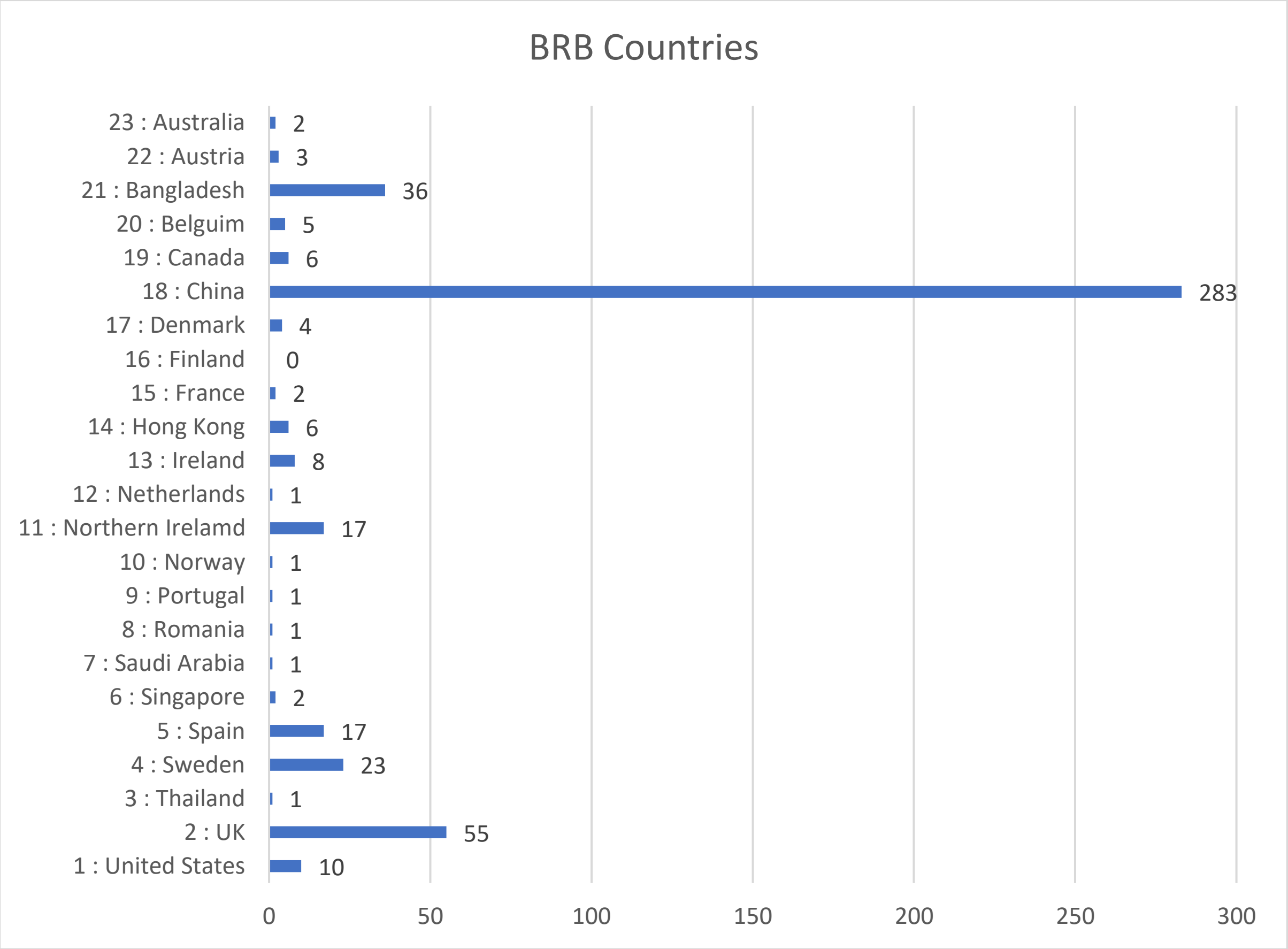

Figure 9

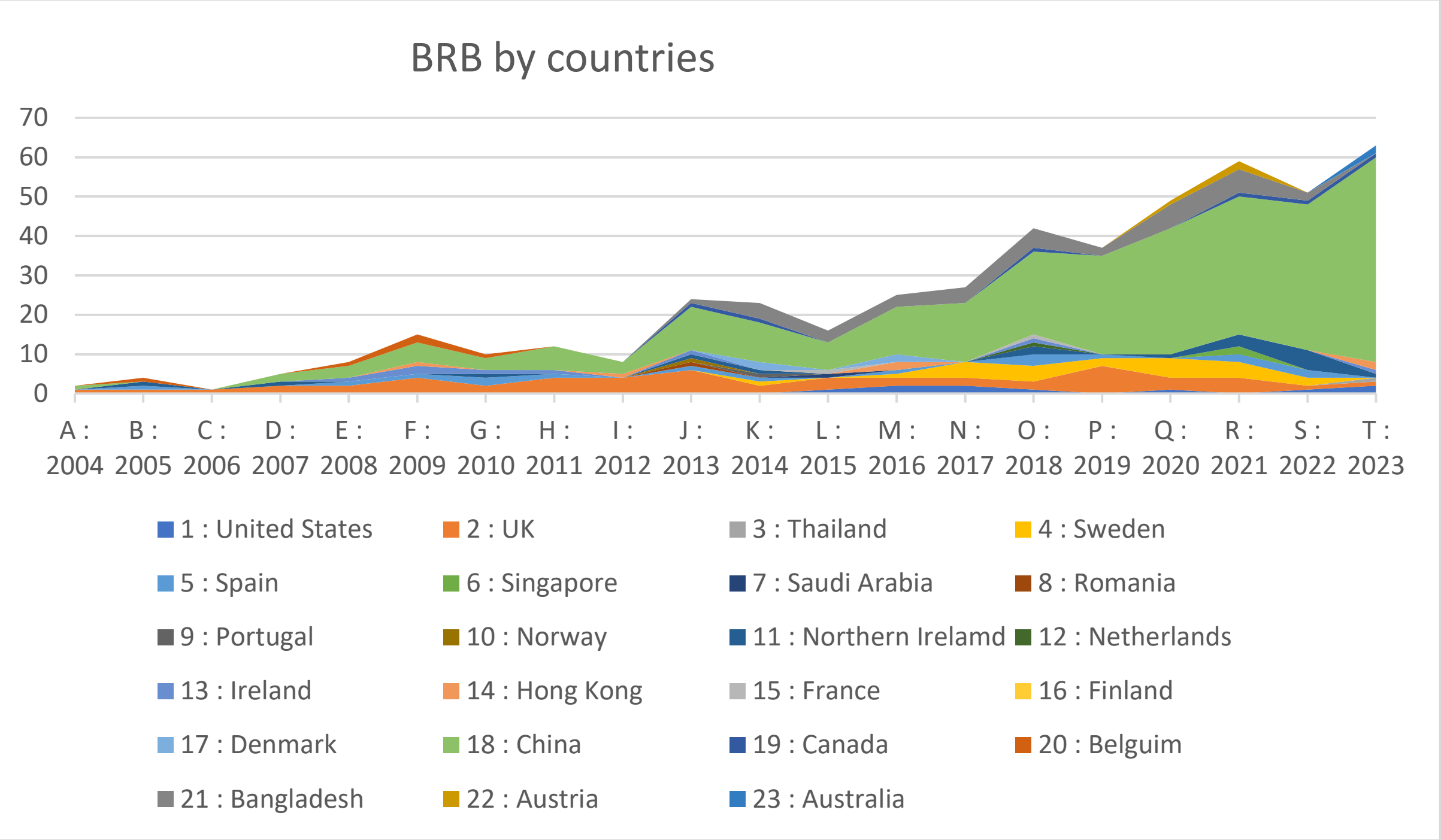

*Figure 10*

# 5 Concluding Remarks

BRB has grown in popularity in recent years and research in recent times has been focussed on practical applications of the technique over a growing number of domains that have been dominated by engineering and industrial. Despite how well suited the technique is to data sets that share features with both insurance and legal in respect of uncertainty, incompleteness and inconsistency, no papers to date have explored the potential. The growing number of papers exploring the application of the technique to stock price movement points to the potential across financial technologies. The new approaches to using large language models to extract rules and attributes from text also point to the potential for combining the explainability of BRB with the raw text processing power of contemporary language models. The next step in our research will be to apply ER / BRB in insurance claims, where insurance and legal expertise intersect.

- The author of this report is an employee of Kennedys Law LLP which is developing products related to the research in this paper.